\pdfoutput=1

\documentclass[11pt]{article}

\usepackage[preprint]{acl}

\usepackage{times}
\usepackage{latexsym}
\usepackage{booktabs}
\usepackage{multirow}

\usepackage[T1]{fontenc}

\usepackage[utf8]{inputenc}

\usepackage{microtype}

\usepackage{inconsolata}

\usepackage{graphicx}

\title{Automatic Generation of Inference Making Questions for Reading Comprehension Assessments}

\author{Wanjing Anya Ma \\
   Stanford University \thanks{Work done while at ETS Research Institute} \\ Stanford, CA, USA\\
  \texttt{wanjingm@stanford.edu}\\\And
  Michael Flor \\
  ETS Research Institute \\ Princeton, NJ, USA \\
  \texttt{MFlor@ets.org}\\\And 
  Zuowei Wang \\
  ETS Research Institute \\ Princeton, NJ, USA \\
  \texttt{zwang@ets.org}\\
  }

\begin{document}
\maketitle
\begin{abstract}
Inference making is an essential but complex skill in reading comprehension (RC). Some inferences require resolving references across sentences, and some rely on using prior knowledge to fill in the detail that is not explicitly written in the text. Diagnostic RC questions can help educators provide more effective and targeted reading instruction and interventions for school-age students. We introduce a taxonomy of inference types for RC and use it to analyze the distribution of items within a diagnostic RC item bank. Next, we present experiments using GPT-4o to generate bridging-inference RC items for given reading passages via few-shot prompting, comparing conditions with and without chain-of-thought prompts. Generated items were evaluated on three aspects: overall item quality, appropriate inference type, and LLM reasoning, achieving high inter-rater agreements above 0.90. Our results show that GPT-4o produced 93.8\% good-quality questions suitable for operational use in grade 3-12 contexts; however, only 42.6\% of the generated questions accurately matched the targeted inference type. We conclude that combining automatic item generation with human judgment offers a promising path toward scalable, high-quality diagnostic RC assessments.

\end{abstract}

\section{Introduction}
\begin{figure*}[ht]
  \centering
  \includegraphics[width=0.9\linewidth]{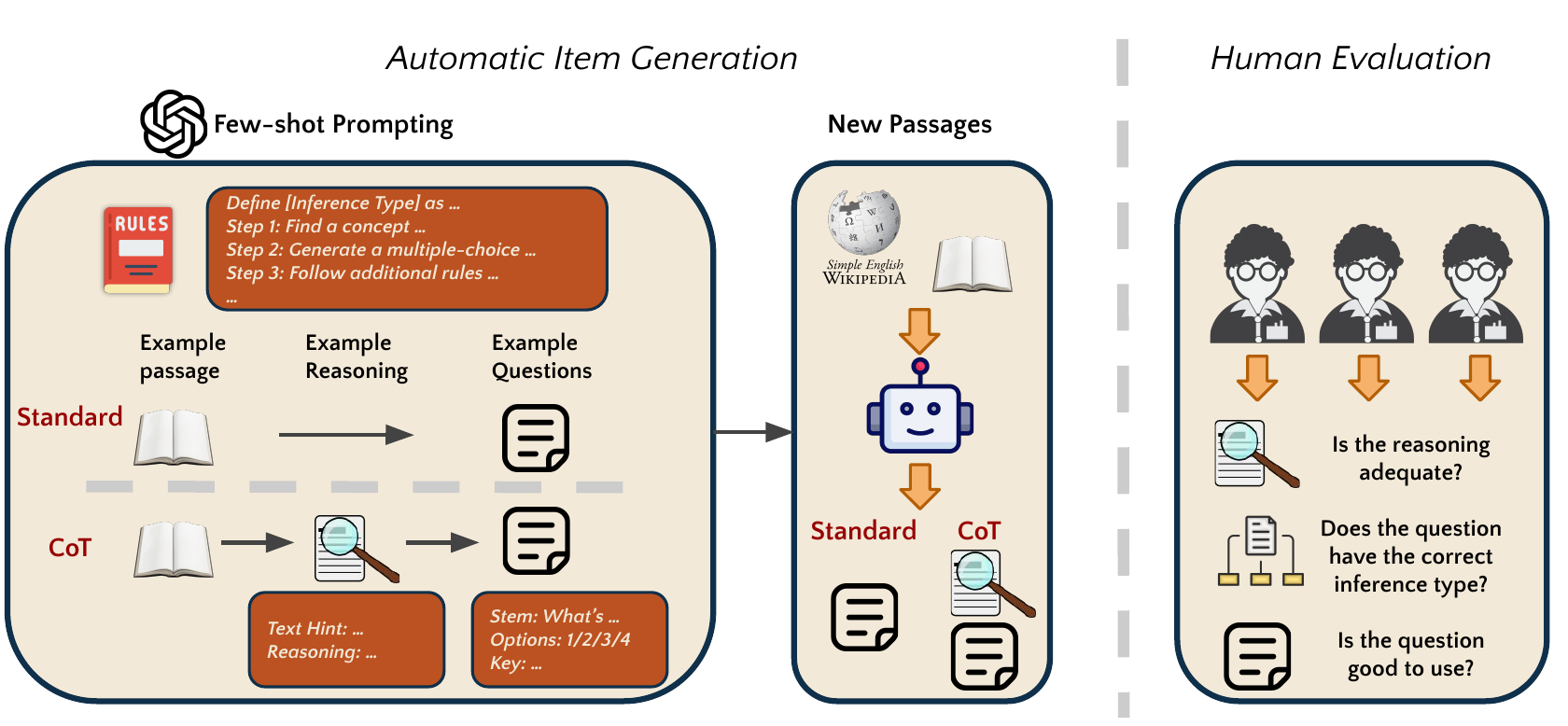} 
  \caption {\textbf{Overview of automatic item generation and human evaluation.} We use GPT-4o to generate bridging-inference RC items for given reading passages via few-shot prompting, comparing conditions with and without chain-of-thought prompts. We prompt each inference type separately: pronominal bridging, text-connecting, and gap-filling inferences. Human evaluation focuses on general item quality, inference type appropriateness, and LLM rationales.}
  \label{fig:overview}
\end{figure*}

Inference-making is an essential yet cognitively demanding skill in reading comprehension (RC) \citep{o2015inferences, kintsch1998comprehension}. Inferences are necessary for establishing both local and global coherence within the mental representation of a text \citep{graesser1994constructing}. Local inferences connect information across sentences using cohesive devices such as anaphors or category exemplars—for example, in "Bette gulped down the drink. The cold water was very refreshing," the reader infers that \textit{the drink} refers to \textit{cold water} \citep[p.~307]{cain2022children}. Global inferences, on the other hand, rely on the reader's prior knowledge to fill in missing details required to make sense of the text —for example, in "The campfire started to burn uncontrollably. Tom grabbed a bucket of water" \citep[p.~192]{bowyer2005assessing}, the reader infers that Tom intended to put out the fire, based on the knowledge that water extinguishes fire. While skilled readers often generate inferences automatically as they engage with text \citep{thurlow1997automaticity}, children who struggle with comprehension frequently have difficulty constructing these inferences \citep{cain2001comprehension}.

Providing diagnostic information about specific types of inference-making deficits that hinder comprehension can empower educators to provide more effective and targeted reading instruction and intervention \citep{bowyer2005assessing, bayat2020relationship}. To achieve this, we need RC assessments that specifically target inference-making types. At the same time, we want to develop scalable item generation methods to enable multi-time testing, monitoring reading development over time. Previous work has demonstrated the ability of large language models (LLMs) to generate effective RC questions \citep{uto-etal-2023-difficulty, sauberli-clematide-2024-automatic}. However, whether LLMs can reliably produce questions that target specific inference types remains unclear. 

Our research is grounded in a real-world diagnostic assessment of reading skills for students in grades 3 through 12 \citep{sabatini2019sara}. The assessment was originally developed at ETS and recently commercialized as ReadBasix. It leverages the science of reading to assess foundational reading skills, such as word recognition and decoding, as well as more complex ones such as RC. In the RC subtest, a student will usually read 4 expository passages and answer multiple-choice questions associated with the passages. The subtest takes about 30 minutes to complete. Like any large-scale reading assessment, there is an ongoing need for more items. To address this demand, we aim to leverage automatic item generation to create new items based on curated passages, and evaluate the quality of these items before collecting student performance data to make them operational.  

For the purpose of automatic item generation, as illustrated in Figure~\ref{fig:overview}, we first conducted a literature review on inference-making in the reading comprehension and natural language processing (NLP) text comprehension literature. We developed a taxonomy of inference-making questions, with a focus on bridging inference. We validated this taxonomy by annotating an operational item bank of expert-written RC questions, confirming bridging inference as an important and widely covered sub-construct. Next, we curated six expository passages and manually wrote multiple-choice RC questions for each inference type based on our taxonomy. These examples were then used to prompt GPT-4o \citep{hurst2024gpt} via few-shot prompting to generate bridging-inference questions for new reading passages, comparing conditions with and without chain-of-thought (CoT) prompting \citep{wei2022chain}. Finally, three human experts evaluated the quality of the generated questions along three dimensions: overall item quality \footnote{The evaluation of overall item quality does not include whether an item is of the required inference type, which is an extra-evaluation. See Table~\ref{tab:eval_rubrics} for more details.}, appropriate inference type, and whether GPT-4o provided satisfactory reasoning for generating the question. Our results show that LLMs can produce 93.8\% good-quality questions suitable for operational use in grade 3-12 contexts; however, only 42.6\% of the generated questions accurately match the targeted inference type. Nevertheless, the overall coverage of inference types closely mirrors what we observe in our operational item bank. We conclude that combining automatic item generation with human judgment offers a promising path toward scalable, high-quality diagnostic RC assessments. 
 
In summary, we make the following contributions in this paper:
\begin{enumerate}
    \item We develop and validate a taxonomy for inference-making questions used in multiple-choice RC assessments, and demonstrate its its value for future item development.
    \item We introduce a novel NLP task where language models generate RC questions targeting specific inference types, providing a new way to assess their reasoning abilities. The training item bank will be released for replication and benchmarking.
    \item We demonstrate GPT-4o's potential in generating RC questions for operational use and its limitations in accurately generating specific types of inference questions. 
\end{enumerate}

\section{Related Work}
\subsection{Question generation for reading comprehension assessments} Automatic question generation is a well-established task in NLP, especially within educational applications, to reduce the high costs of manual question authoring and to ensure a steady supply of new, high-quality items \citep{kurdi2020systematic}. Early approaches rely on rule-based or template-based methods \citep{araki2016generating, flor2018semantic}, as well as the use of discourse connectives to generate questions \citep{agarwal2011automatic}. Later approaches extensively used neural systems for question generation \citep{MullaGharpure2023}. More recent work demonstrates that LLMs hold promise in generating high-quality RC questions, using techniques such as fine-tuning \citep{uto-etal-2023-difficulty, perkoff2023comparing, ghanem-etal-2022-question, ashok-kumar-etal-2023-improving, rathod-etal-2022-educational, stasaski2021automatically}, zero-shot or few-shot prompting \citep{sauberli-clematide-2024-automatic, attali2022interactive}, and Chain-of-Thought prompting \citep{kulshreshtha2022reasoning}. Some of these studies have also explored the generation of more complex, "deeper" questions—those that target underlying reasoning processes \citep{ghanem-etal-2022-question, poon-etal-2024-shot} or hinge on specific inference steps for accurate responses \citep{araki2016generating}. Within the domain of automated Question Answering, the notion of 
\textit{multi-hop questions} has gained attention, as questions relating different parts of a document require multi-step reasoning
\cite{mavi2024multihopquestionanswering}. 

We note that prior studies have largely treated reading comprehension as a single, undifferentiated construct even though comprehension requires different types of inferences. Recent work has begun to develop taxonomies of RC and annotate question types to enable more controllable generation \cite{xu2022fantastic, li2024planning, hwang2024towards}. However, to our knowledge, no existing work has systematically addressed question generation based on specific types of inference. We believe that the capability to generate different types of inference questions will provide more diagnostic insights for educators. Our work is a first step toward filling this gap. 

\subsection{Bridging inference as an NLP task} The NLP community has long tackled text comprehension challenges, including bridging inference. Prior work has focused on corpus-based bridging anaphora recognition and resolution using annotated resources such as ISNotes and BASHI \citep{rosiger-2018-bashi, hou2018unrestricted, hou-2020-bridging}. Neural models have been developed to jointly learn mention representations and bridging relations \citep{pandit-hou-2021-probing, kobayashi-etal-2022-end}. In the recently developed IdentifyMe benchmark for resolving nominal and pronominal mentions across long contexts \citep{manikantan2024identifyme}, GPT-4o outperforms other LLMs, achieving 81.9\% accuracy and demonstrating strong referential capabilities. With the rise of LLMs, research increasingly shifts toward evaluating LLMs' general reasoning capabilities \citep{brown2020language,wei2022chain}. In our education application, we investigate whether LLMs truly possess the reasoning ability required for bridging inference, particularly through the lens of a question generation task.

\section{Taxonomy of Inference Questions}
\subsection{Development of Taxonomy}
\begin{table*}[ht!]
\centering
\resizebox{\linewidth}{!}{
\begin{tabular} {p{0.11\linewidth} p{0.25\linewidth} p{0.64\linewidth}}
\toprule
\textbf{Types} & \textbf{Definitions} & \textbf{Examples} \\
\midrule
Pronominal & Direct pronoun resolution. & Like \textit{"To whom `he' refers?", "What does `this' represent?"} \\
\addlinespace
Pronominal Bridging & Use pronoun as a hint to bridge sentences. & 
Text snippet: \textit{Ships have carried passengers since prehistoric times. That is the first kind of public transportation.} \newline
Question: \textit{What was the first kind of public transportation in history?} \newline
Answer: \textit{ships} \newline
Reasoning: \textit{The pronoun "That" refers to "ships" in the previous sentence.} \\
\addlinespace
Text-Connecting & Connecting two explicitly stated components in a text, typically through a noun phrase. &
Text snippet: \textit{Public transportation is good for the environment. When many people use the same vehicle, fewer cars are on the road. Fewer cars make less pollution.} \newline
Question: \textit{Why is public transportation good for the environment?} \newline
Answer: \textit{Because it causes less pollution} \newline
Reasoning: \textit{ "Fewer cars" links to "public transportation" from the previous sentence in a causal relationship.} \\
\addlinespace
Gap-Filling & "Incorporating information outside of the text, i.e., general knowledge, with information in the text to fill in missing details." \citep[p.490]{cain1999inference} &
Text snippet: \textit{White pizza uses no tomato sauce, often substituting pesto or dairy products such as sour cream. Most commonly, its toppings consist only of mozzarella and ricotta cheese drizzled with olive oil and basil and garlic.} \newline
Question: \textit{What is a possible reason "White pizza" gets its name?} \newline
Answer: \textit{It doesn't have tomato sauce} \newline
Reasoning: \textit{Readers need to use common sense to fill in the gap that "no tomato sauce" means the color of the pizza is not red.} \\

\bottomrule
\end{tabular}
}
\caption{Taxonomy of inferences for Reading Comprehension questions.}
\label{tab:taxonomy}
\end{table*}

Inferences can be categorized into bridging inferences, elaborative inferences, predictive inference, emotional inference, etc \citep{graesser1994constructing, schmalhofer2002unified, singer2004retrieving, van2015inference}. To manage the scope of our interest, we focus on bridging inference which connects information in a text. Bridging inferences contribute to text coherence by allowing the reader to identify the connections among concepts and ideas in the text \citep{ singer1992individual, singer2004retrieving} or bridges \citep{haviland1974s} among the propositions underlying the discourse. A bridging inference is needed when the reader cannot retrieve a referent for the given information of the current sentence from either working memory or long-term memory. 

Table ~\ref{tab:taxonomy} shows the taxonomy of inference making questions for diagnostic RC assessments, along with the examples. The first type is \textbf{pronominal}, and it has two variants. Simple pronominal asks for a direct pronoun resolution, such as "In the sentence, whom does `he' refer to?" This is different from the second subtype: \textbf{pronominal bridging}, which requires the reader to use the pronoun as a hint to bridge sentences and answer the question. The third type \textbf{text-connecting} requires test takers to connect two explicitly stated components in a text, and usually the bridge are noun phrases. The last type is \textbf{gap-filling}, which requires readers to incorporate information from outside of the text with information in the text to fill in some missing details. More examples based on the taxonomy are included in Appendix~\ref{bridging inference examples}. 

\subsection{Validation of Taxonomy}
With the newly developed taxonomy, we annotated the RC items in an in-house item-bank. The item-bank has 192 expert-written multiple-choice RC questions for 24 expository reading passages. These passages vary in difficulty from Grade 3 to Grade 12. Our primary focus was to classify the types of bridging inferences, but we also annotated questions that are not in our main scope of interest. For example, there are some \textbf{factual/literal} questions, for which a test taker can directly find information from the text without involving inference; \textbf{vocabulary} questions that directly assess the vocabulary knowledge, and other comprehension questions that do not require bridging inferences. 

\begin{figure}[t]
  \includegraphics[width=\columnwidth]{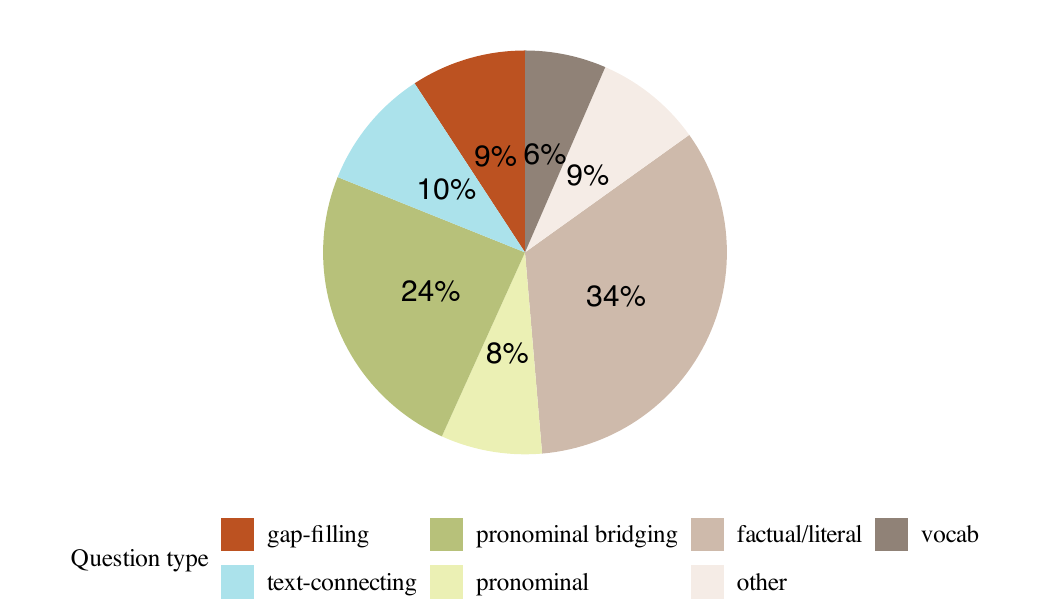}
  \caption{Distribution of different inference types in an operational reading comprehension item bank.}
  \label{fig:operational inference coverage}
\end{figure}

Two of the co-authors classified items independently, following the annotation guideline (see Appendix~\ref{Annotation Guidelines}). The two coders provided the same coding of the type of inference on 86\% of the items, with kappa = $0.83$, indicating high agreement. Based on our annotation results shown in Figure~\ref{fig:operational inference coverage}, we find that bridging inference questions account for 51\% of the RC items in the item bank, suggesting bridging inference is an important sub-construct in this RC assessment. Among the bridging inference questions, pronominal bridging (24\%) is the most dominant type, followed by text-connecting (10\%), gap-filling (9\%), and pronominal questions (8\%). The high level of agreement supports the validity of the newly developed taxonomy, which we see as an important contribution— providing a road-map for both item development and future research. 

\section{Automatic Item Generation and Human Evaluation}
Figure~\ref{fig:overview} presents the overview of our automatic item generation pipeline. 

\subsection{Training Questions}
Due to test security considerations we can not use texts and items from our operational item bank as examples to prompt LLMs. Thus, we created our example item bank which is publicly available for replication efforts \footnote{\url{https://github.com/maafiah/InferenceQuestionsAQG}}. We adapted 6 new expository passages from Simple English Wikipedia \footnote{\url{https://simple.wikiedia.org}} (passage length ranges from 342 to 508 words, average 438) and for each passage we manually created 2-4 items for each type of inference. Each question contains a \textbf{stem}, four \textbf{options}, and an answer \textbf{key} indicating which option is correct. We also included our thought process in the item generation: \textbf{text hint} includes the relevant text from the passage where required inference will be made, and \textbf{reasoning} is a short explanation why this question belongs to the requested type. In total, we wrote 19 pronominal bridging, 23 gap-filling, and 16 text-connecting questions. 

\subsection{Few-shot Prompting}
\begin{figure}[ht!]
  \includegraphics[width=\columnwidth]{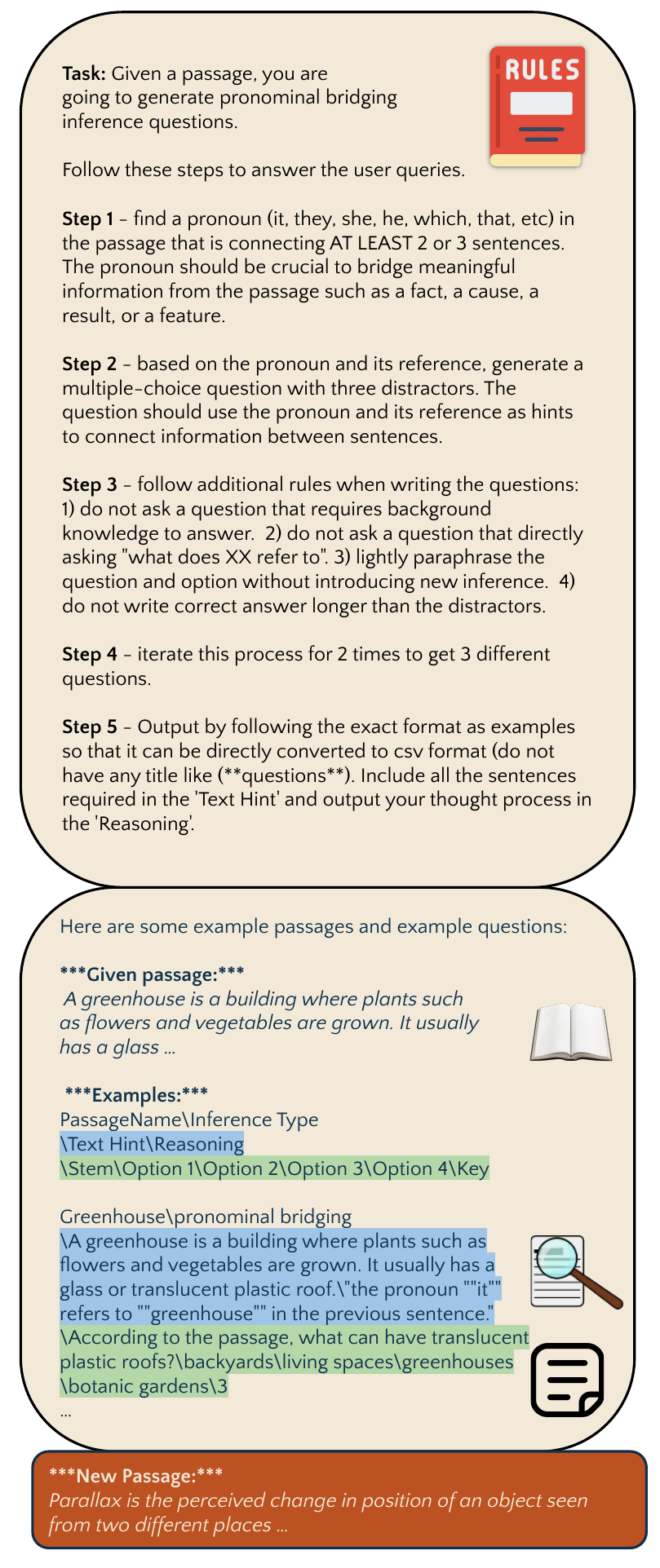}
  \caption{\textbf{Few-shot prompt for generating pronominal bridging inference questions.} The system prompt (beige background) defines the inference type and outlines expert-inspired steps. Training examples (provided in the prompt) follow. In the standard condition, only the question and answer key (green) are shown; in the CoT condition, text hints and reasoning (blue) are also included. A new passage is provided in the user prompt (orange background) to generate new questions.}
  \label{fig:prompting_demo}
\end{figure}

\renewcommand{\arraystretch}{2} 
\begin{table*}[h]
  \centering
  \small
  \resizebox{\linewidth}{!}{
\begin{tabular}{p{0.2\linewidth} p{0.8\linewidth}}
    \hline
    \textbf{Criterion} & \textbf{Annotation Guidelines} \\
    \hline
    General item quality &
    1: If the generated item satisfies all of the following: \par
    (a) The correct answer is fully correct; \par
    (b) Distractors are not confusing and are clearly incorrect; \par
    (c) The question is developmentally appropriate and safe for Grades 3–12. \par
    0: If any requirement is not met. Provide an explanation in the "Note" field. \\
    
    Inference-type accuracy &
    1: If the generated item matches the requested inference type. \par
    0: If not. \newline 
    Output inference type, one of: \newline gap-filling / pronominal bridging / text-connecting / factual or literal. \\

    Reasoning quality &
    1: If the generated thought process fulfills both of the following: \par
    (a) The "Reasoning" is adequate and relevant to the requested inference type; \par
    (b) The "Text Hint" includes all the sentences required to answer the item correctly. \par
    0: If either condition is not satisfied. \\
    \hline
  \end{tabular}
  }
  \caption{\label{tab:eval_rubrics}
    Annotation guidelines for evaluating the generated items.
  }
\end{table*}

We used the GPT-4o model (2024-04-01-preview) to generate multiple-choice RC questions based on passages we supplied to the model. To prioritize accuracy and reproducibility in item generation, we set the temperature parameter to 0. We explored the frequency penalty parameter from 0 to 0.3, with 0.2 proving optimal as it could consistently generate three diverse RC items without compromising their quality.

Few-shot prompting techniques were used and the prompts were iteratively refined over six rounds. Most adjustments focused on improving the concreteness of the question-writing steps to better guide the model. In this paper, we only report the final iteration of item generation in which we experimented with four different prompting conditions: standard prompting with 4 (or 6) passages and examples, and chain-of-thought prompting with 4 (or 6) passages and examples with text hint and reasoning. With this set-up, we investigated whether increasing the training examples or using the CoT strategy would improve the quality of generation. Moreover, we further evaluated if the output reasoning process was adequate for this specific task. 

Figure~\ref{fig:prompting_demo} shows an example prompt for generating reading comprehension questions targeting pronominal bridging inference (see Appendix~\ref{Prompting} for more details). In the system prompt, we first instructed GPT-4o to identify pronominal bridging relationships, then directed it to generate a multiple-choice question, guided by additional rules to ensure item quality. We included several training examples in the prompt —either 4 or 6 passages with corresponding questions, depending on the generation condition. For the Standard condition, no text hints or reasoning were provided in the training examples. In the CoT condition, both text hints and reasoning were provided, prompting the model to generate them in the output. In the user prompt, we provided a new passage for GTP-4o to generate items from. 

We curated a total of 10 new passages adapted from Simple Wikipedia, which were comparable in length and format to the example passages. For each passage and inference type (pronominal bridging, text-connecting, and gap-filling), we independently applied the prompting procedure, instructing GPT-4o to generate three unique questions per combination. For text-connecting and gap-filling—where question construction can be more challenging—we included an additional rule: "Do not force additional questions if no suitable locations can be found." Across the four prompting conditions, we generated a total of 357 questions, 180 of which were produced under the CoT condition and therefore included text hints and reasoning in the output.

\subsection{Human Evaluation}
To evaluate the quality of the generated RC items, we developed an evaluation rubric (see Table~\ref{tab:eval_rubrics}). Three authors used items from prior iterations of the generation process and complete several practice rounds and discussion before finalizing the rubric. The rubric is designed to directly address our core research questions:

\begin{description}
  \item[RQ1:] Can LLMs generate high-quality RC items with appropriate distractors suitable for inclusion in an operational item bank?
  \item[RQ2:] Do the generated RC items align with the requested bridging inference type?
  \item[RQ3:] How well can LLMs reason about their generation process?
\end{description}

In the evaluation phase, the three authors, who are experts in reading assessment questions, independently annotated all 357 generated items. The agreement was high for general item quality (RQ1), with percent agreement ranging from 87–90\%. However, reaching consensus on the inference type (RQ2; 69–70\%) and reasoning quality (RQ3; 65–71\%) proved more challenging—consistent with prior findings that reasoning-related judgments are inherently difficult to rate \citep{stasaski2021automatically}.

\begin{table}[t]
  \centering
  \small
  \begin{tabular}{lrr}
    \hline
    \textbf{Criterion} & \textbf{Agreement (\%)} & \textbf{Fleiss' $\kappa$} \\
    \hline
    General item quality  & 90–97 & 0.57 \\
    Inference-type accuracy     & 85–94 & 0.77 \\
    Reasoning quality      & 90–95 & 0.83 \\
    \hline
  \end{tabular}
  \caption{\label{tab:agreement}
    Inter-rater agreement and Fleiss' $\kappa$ for each evaluation criterion. Agreement is reported as a range based on three pairwise comparisons by three graders.}
\end{table}

\begin{figure*}[t]
  \includegraphics[width=\linewidth]{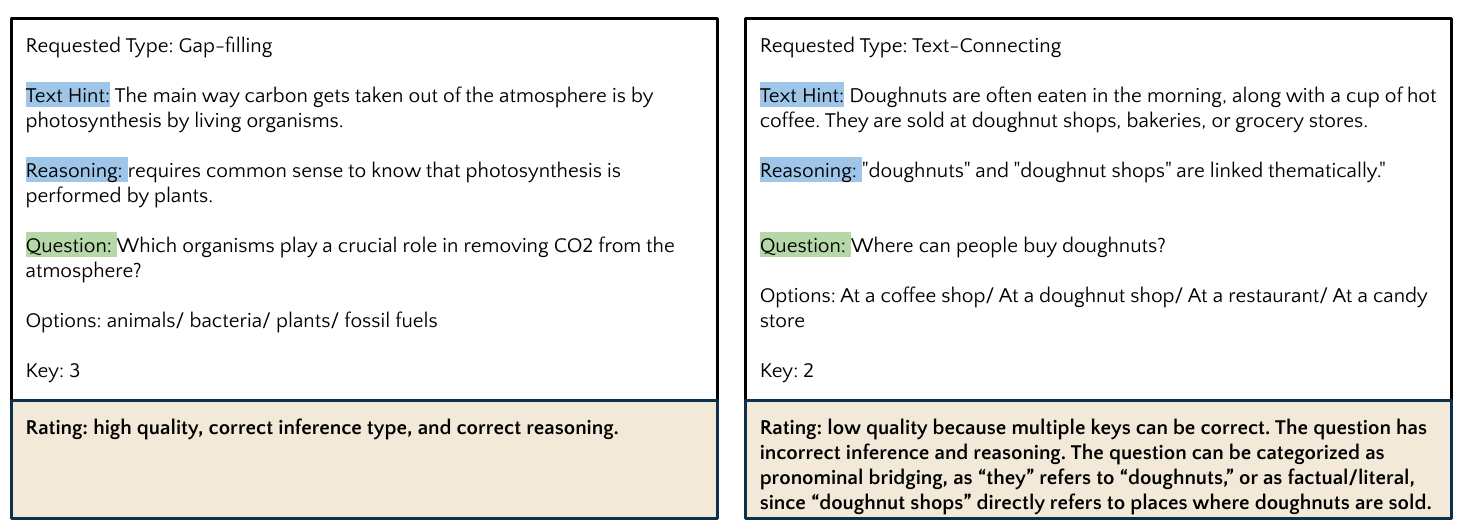}
  \caption{Examples of LLM-generated RC items via Chain-of-Thought prompting with 6 training passages. Left: high-quality; right: low-quality. Each output includes a text hint, a rationale, a multiple-choice question with four options, and an answer key. Human annotations are shown against a beige background.}
  \label{fig:AIG_question_demo}
\end{figure*}

To address this, we conducted a second round of annotation. In this phase, each rater independently reviewed only the items where their initial rating differed from the other two and decided whether to adjust the rater's original score. Following this adjustment, inter-rater agreement improved substantially. The final results of percentage agreement and Fleiss' kappa are shown in Table~\ref{tab:agreement}. Our evaluation in the Results section were based on the majority votes for each item. For example, an item was treated as acceptable when at least two of the three raters rated it as good quality.

\section{Results}
Based on the proportion of accepted items by generation method (Table~\ref{tab:result_table}), we observe improved generation performance when increasing the number of training examples from four to six example passages in the prompt. However, our experiment does not show any clear advantage of Chain-of-Thought prompting over standard few-shot prompting. Furthermore, our results indicate no statistically significant differences in generation performance across the various prompting conditions. We summarize our key findings below.

\vspace{6px}
\noindent\textbf{LLMs can produce high-quality questions suitable for operational use.} Based on the evaluation of general item quality, 87 out of 90 questions (96.7\% in the CoT\_6 condition) had good quality and were suitable for operational use in the Grade 3-12 educational context. The performance is comparable to, if not better than, those reported in prior research evaluating overall item quality for RC assessments, which ranged from 75\% to 90\% \citep{kulshreshtha2022reasoning, uto-etal-2023-difficulty, sauberli-clematide-2024-automatic}. Because of the differences between these studies, for a more informative comparison, we encourage future research to replicate our findings under similar conditions. Figure~\ref{fig:AIG_question_demo} presents one high-quality example and one low-quality example of the generated questions. We find that problems of unacceptable questions included multiple keys, introduction of new vocabulary, confusing wording of the question, etc. 

\begin{table}[]
\small
\resizebox{\columnwidth}{!}{%
\begin{tabular}{ccccc}
\hline
\textbf{\begin{tabular}[c]{@{}c@{}}Generation\\ Method\end{tabular}} &
  \textbf{\begin{tabular}[c]{@{}c@{}}Num \\ Items\end{tabular}} &
  \textbf{\begin{tabular}[c]{@{}c@{}}General \\ Item Quality\end{tabular}} &
  \textbf{\begin{tabular}[c]{@{}c@{}}Inference \\ Accuracy\end{tabular}} &
  \textbf{\begin{tabular}[c]{@{}c@{}}Reasoning \\ Quality\end{tabular}} \\
  \hline
  standard\_4 & 88 & 0.932 & 0.409 & \   \\
standard\_6 & 89 & 0.955 & \textbf{0.461} & \ \\
CoT\_4      & 90 & 0.900 & 0.411 & 0.356 \\
CoT\_6      & 90 & \textbf{0.967} & 0.422 & \textbf{0.389} \\
\hline
Total      & 357 & 0.938 & 0.426 & 0.372 \\
\hline
\end{tabular} 
}
\caption{\label{tab:result_table}
    Proportion of accepted items by generation method—standard vs. chain-of-thought prompting (with text hints and reasoning), using 4 or 6 passages (12–18 examples). Highest scores per criterion are bolded; criteria are defined in Table~\ref{tab:eval_rubrics}.
  }
\end{table}

\begin{figure*}[t]
\centering
  \includegraphics[width=0.8\linewidth]{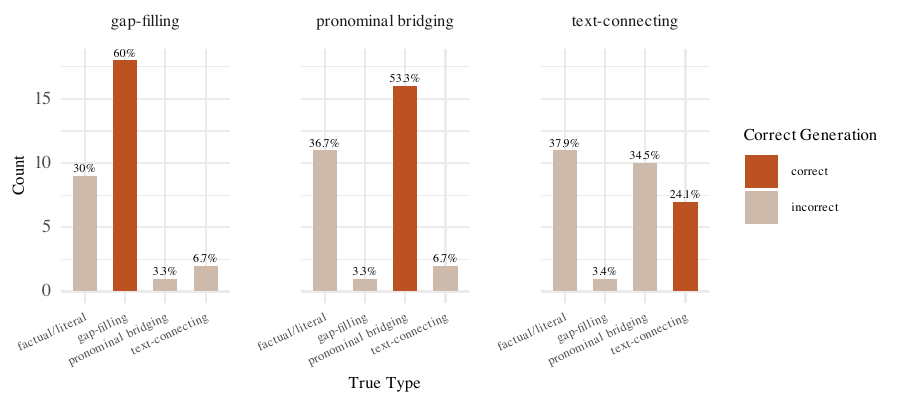}
  \caption{Human evaluation of inference-type accuracy. Each panel displays the distribution of true inference types corresponding to each requested inference type. The generation questions are obtained from the standard few-shot prompting with 6 training passages.  }
  \label{fig:inference_eval}
\end{figure*}

\begin{figure}[h]
  \includegraphics[width=1\columnwidth]{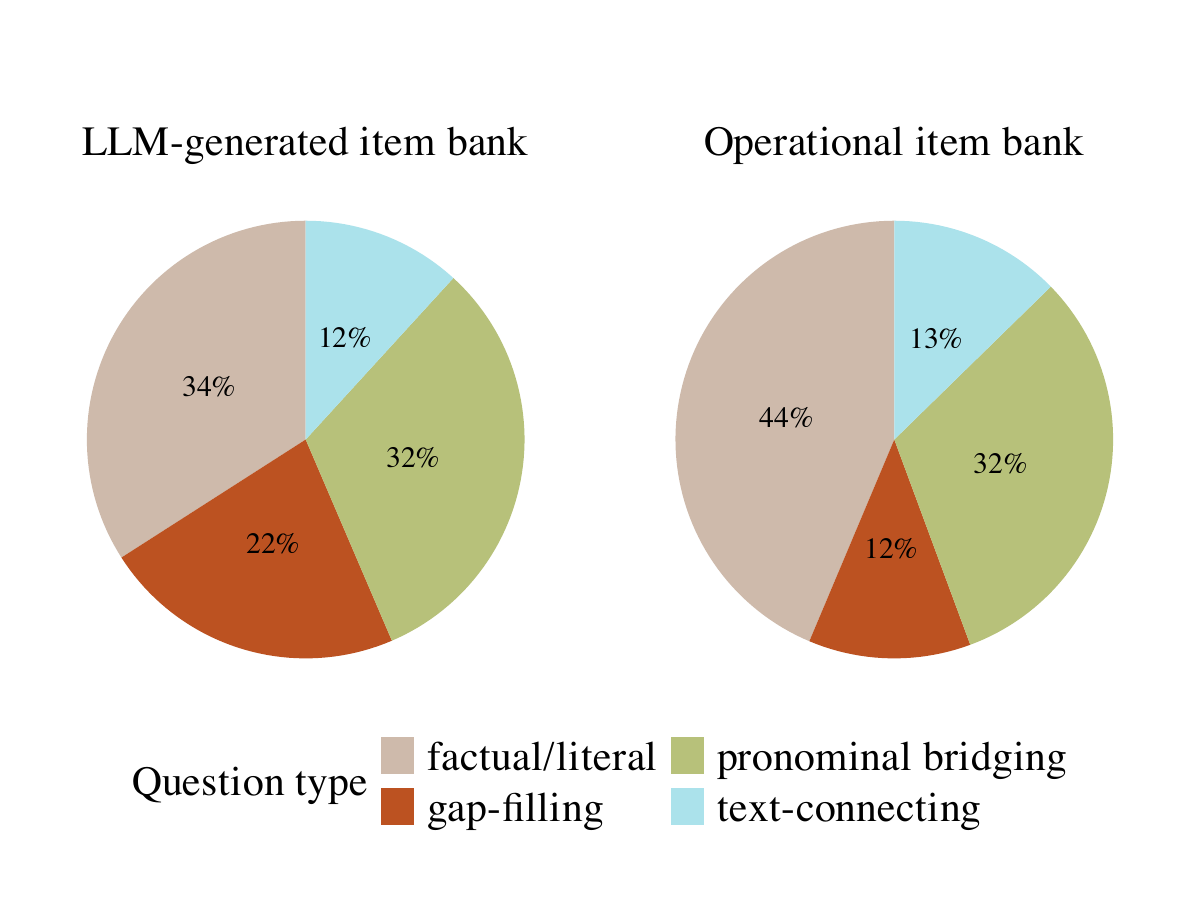}
  \centering
  \caption{Comparison of item inference type coverage between the operational item bank and the LLM-generated item bank.}
  \label{fig:coverage_compare}
\end{figure}

\vspace{6px}
\noindent\textbf{Generating RC questions by specific inference type is a challenging NLP task.} Although LLMs can generate high-quality RC items, their ability to produce questions targeting specific inference types remains limited. In the generation method yielding the best performance (standard\_6), only 46.1\% of the generated questions matched the requested inference type. As shown in Figure~\ref{fig:inference_eval}, gap-filling questions were the easiest to generate (60\% match), followed by pronominal-bridging questions (53.3\% match). In contrast, generating text-connecting questions proved particularly difficult, with an accuracy of only 24.1\%. This pattern of generation difficulty aligns with the challenges faced by human experts (co-authors) when writing the training examples. We also find that 34.8\% of the generated questions were factual or literal, requiring little inference. Moreover, GPT-4o provided adequate reasoning for only 38.9\% of the items. This finding may explain the lack of performance gains when moving from standard prompting to CoT prompting. While prior work has shown that adding structured rationales can improve the accuracy of multi-hop question generation \citep{sauberli-clematide-2024-automatic}, we believe our task poses a more challenging test of an LLM's reasoning ability.

\vspace{6px}
\noindent\textbf{Automatic item generation with human evaluation ensures the quality of diagnostic RC items.}
From an application standpoint, we also examined how closely the distribution of inference types in the generated items resembled that of human-written items from our operational RC item bank. Interestingly, our analysis, shown in Figure ~\ref{fig:coverage_compare}, reveals that the overall distribution of inference types in the LLM-generated items closely matches that of our operational RC item bank. This means whereas GPT-4o failed to consistently produce individual items targeting specific inference types, the collect of items it generated somehow resembles the distribution of item types in our existing item pool. With some expert review, most of these items are suitable to use. Understanding the strengths and limitations of current LLM performance is important, particularly if we aim to rely on human evaluation to ensure quality and safety. The generation process is considerably more scalable than relying on human experts to write items manually. Despite current limitations, LLM-based item generation with our newly developed taxonomy offers a promising approach for educational applications.

\section{Conclusion}
This paper demonstrates our effort in leveraging a large language model to generate inference-making questions for a reading comprehension assessment. We developed a taxonomy of bridging inference questions based on existing literature and validated it with empirical data from an operational test. The taxonomy focuses on three types of inferences: pronominal bridging, text connecting, and gap-filling. The taxonomy guided our manual creation of example comprehension questions, which were then used as training materials for GPT-4o to generate new items for the new passages. Our evaluation indicates that although GPT-4o can produce acceptable RC questions, its ability to generate questions aligned with specific inference types was limited. This limitation might stem from its limited capability in providing valid reasoning for the types of inferences. These results highlight the critical role of human evaluation when using LLMs for RC question creation. We propose that combining automatic item generation with human judgment offers a promising path toward scalable, high-quality diagnostic RC assessments. 

\section*{Limitations}

We provide preliminary evidence for the potential of GPT-4o in creating inference making reading comprehension questions. The following limitations should be addressed by future research.

\vspace{6px}
\noindent\textbf{We have a limited evaluation set.}
Our evaluation relies on 10 expository passages (based on Simple Wikipedia), restricting the generalizability of our findings to broader reading contexts or varied educational materials. Future research should incorporate more passages and of different genres, such as narratives. 

\vspace{6px}
\noindent\textbf{We exclusively use GPT-4o.}
This study employed only one LLM, GPT-4o, which may limit insights into the potential effectiveness of other advanced reasoning models. Given the challenge of this reasoning task, future research should explore additional models. Because more advanced models may incur significantly higher costs, future research should also consider the balance between performance and affordability for an educational application.

\vspace{6px}
\noindent\textbf{Unclear effectiveness of Chain-of-Thought prompting.}
Our results show that generation quality improves with more example questions. However, our experiment does not show benefits from CoT prompting. This unexpected finding may result from our limited number of training examples. Future studies should expand the training data and possibly utilize large datasets, such as SQuAD \citep{rajpurkar-etal-2016-squad} and FairytaleQA \citep{xu2022fantastic}. Future work should also explore more effective methods for integrating human-experts' rationales into the question generation process and explore how it affects the reasoning performance of LLMs \citep{zelikman2022star}. 

\vspace{6px}
\noindent\textbf{General item quality is a broad metric.}
Our main goal is to generate RC items that target specific inference types, so we grouped other aspects like answer correctness and distractor plausibility under a broad "General Item Quality" metric. Still, there are important dimensions we didn't separate out—like item difficulty and whether it's appropriate for the target population. More specific metrics could help pinpoint where generation errors happen and how inference type and item difficulty might interact.

\vspace{6px}
\noindent\textbf{Future work should focus on item evaluation in real-world deployment.}
Our study did not include pilot testing in real-world settings to evaluate how the generated items perform with actual student responses. Student response data would allow for further examination of item bias, difficulty, and discrimination—critical steps before using the items for student scoring and making valid inferences about their abilities \citep{yeatman2024development}. Using LLM-simulated student responses to evaluate generated items is also an exciting direction that could help reduce—but not replace—the need for traditional item calibration \citep{zelikman-etal-2023-generating, lu2024generative, liu2025leveraging}. 

\section*{Ethics Statement}
Our study goal is to leverage LLMs to develop scalable and effective RC assessments to align with educational practice. We introduce a novel and meaningful NLP task: generating RC questions by inference type. While LLMs show promise for item development, we emphasize the importance of maintaining test security by avoiding training models on operational test items, and by ensuring the safety of content such as developmental appropriateness and the absence of problematic materials. In addition to existing automatic benchmarks, human evaluation by educational experts remains essential for item quality. Though beyond our current scope, we also highlight the need for ongoing monitoring of the generated items to detect scoring biases and ensure fairness in operational use.

\section*{Acknowledgments}
We appreciate the reviewers for their helpful feedback on the manuscript. This study was made possible by the following research grant awarded by the Institute of Education Sciences, U.S. Department of Education, through R305F100005. Opinions, findings, and conclusions in this paper do not necessarily reflect the views of IES or ETS.

\bibliography{acl_latex}

\clearpage
\newpage

\appendix

\section{Bridging Inference Examples}
\label{bridging inference examples}
When we developed the taxonomy of bridging inference, we referred to a sample passage and a list of example questions provided from \citet[p.495]{cain1999inference}. Table \ref{tab:example-questions} presents our analysis of the given questions based on the taxonomy. 

\begin{table*}
  \centering
  \begin{tabular}{@{}p{0.28\linewidth} p{0.68\linewidth}@{}}
    \hline
    \textbf{Reading Passage:} \\
    \hline
    \multicolumn{2}{p{0.96\linewidth}}{ 
    Debbie was going out for the afternoon with her friend Michael. By the time they got there they were very thirsty. Michael got some drink out of his duffel bag and they shared that. The orange juice was very refreshing. Debbie put on her swimming costume, but the water was too cold to paddle in, so they made sandcastles instead.

    They played all afternoon and didn't notice how late it was. Then Debbie spotted the clock on the pier. If she was late for dinner, her parents would be angry. They quickly packed up their things. Debbie changed and wrapped her swimming costume in her towel. She put the bundle in her rucksack. Then they set off for home, pedalling as fast as they could. Debbie was very tired when she got home, but she was just in time for dinner.} \\
    \hline

    \textbf{Question} & \textbf{Annotation} \\
    \hline

    \multicolumn{2}{p{0.96\linewidth}}{\textbf{Literal information}} \\
    Who did Debbie spend the afternoon with? & The answer is in the first sentence. There is a partial paraphrase: "going out for" vs. "spend". \\
    Where was the clock? & The answer is in the second sentence of the second paragraph. \\

    \multicolumn{2}{p{0.96\linewidth}}{\textbf{Text-connecting inference}} \\
    Where did Michael get the orange juice from? & This requires bridging inference: \textit{drink} = \textit{orange\_juice}. This is both a referential and semantic link (hypernym: drink – hyponym: juice). Recognizing this link requires background knowledge and both components are near each other in the text. \\
    Where did Debbie put her towel when she packed up her things? & The answer is in sentences 5–6 of the second paragraph. This involves recognizing a part-whole relationship (towel–bundle), which is an ad-hoc, situational reference. \\

    \multicolumn{2}{p{0.96\linewidth}}{\textbf{Gap-filling inference}} \\
    Where did Debbie and Michael spend the afternoon? & One component (afternoon) is in the text, but the location (the beach) is not. It must be inferred as a plausible missing piece of the situation model. \\
    How did Debbie and Michael travel home? & The text says "set off for home" (a paraphrase of "travel"). The mode of travel is inferred from "pedalled", enriching the situation model. \\

    \hline
  \end{tabular}
  \caption{\label{tab:example-questions}
    Analysis of a reading passage and associated reading comprehension questions with inference annotations. The passage and questions are adapted from \citet[p.495]{cain1999inference}.
  }
\end{table*}

\section{Annotation Guidelines}
\label{Annotation Guidelines}
To validate the newly developed taxonomy of bridging inference questions, we annotated an in-house RC item bank. Annotation was done with regards to the text and the questions including stem, key and distractors (see details in Table ~\ref{tab:annotation-guideline}). 

\begin{table*}[t]
\centering
\begin{tabular}{p{0.15\linewidth} p{0.15\linewidth} p{0.55\linewidth}}
\hline
\textbf{Dimension} & \textbf{Options} & \textbf{Note} \\
\hline

\multirow{7}{=}{Inference} 
  & Factual / Literal & The answer is explicitly stated in the text, exactly matching the question. No inference needed. \\
  & Pronominal & Resolving pronouns (e.g., "Who does `he' refer to?"). \\
  & Pronominal Bridging & Requires resolving a pronoun and using it as a cue to infer the correct answer. \\
  & Text-Connecting & Requires connecting two explicitly stated components, typically using noun phrases. \\
  & Gap-Filling & Involves filling in a missing but easily inferred piece of information not directly stated in the text. \\
  & Vocabulary & Tests the reader's knowledge of word meanings. \\
  & Other & Any other type, such as comparison or author intent. \\

\hline
\end{tabular}
\caption{\label{tab:annotation-guideline}
Annotation guidelines for the in-house item bank.}
\end{table*}

\section{Prompts}
\label{Prompting}
\begin{figure*}[h]
  \includegraphics[width=\linewidth]{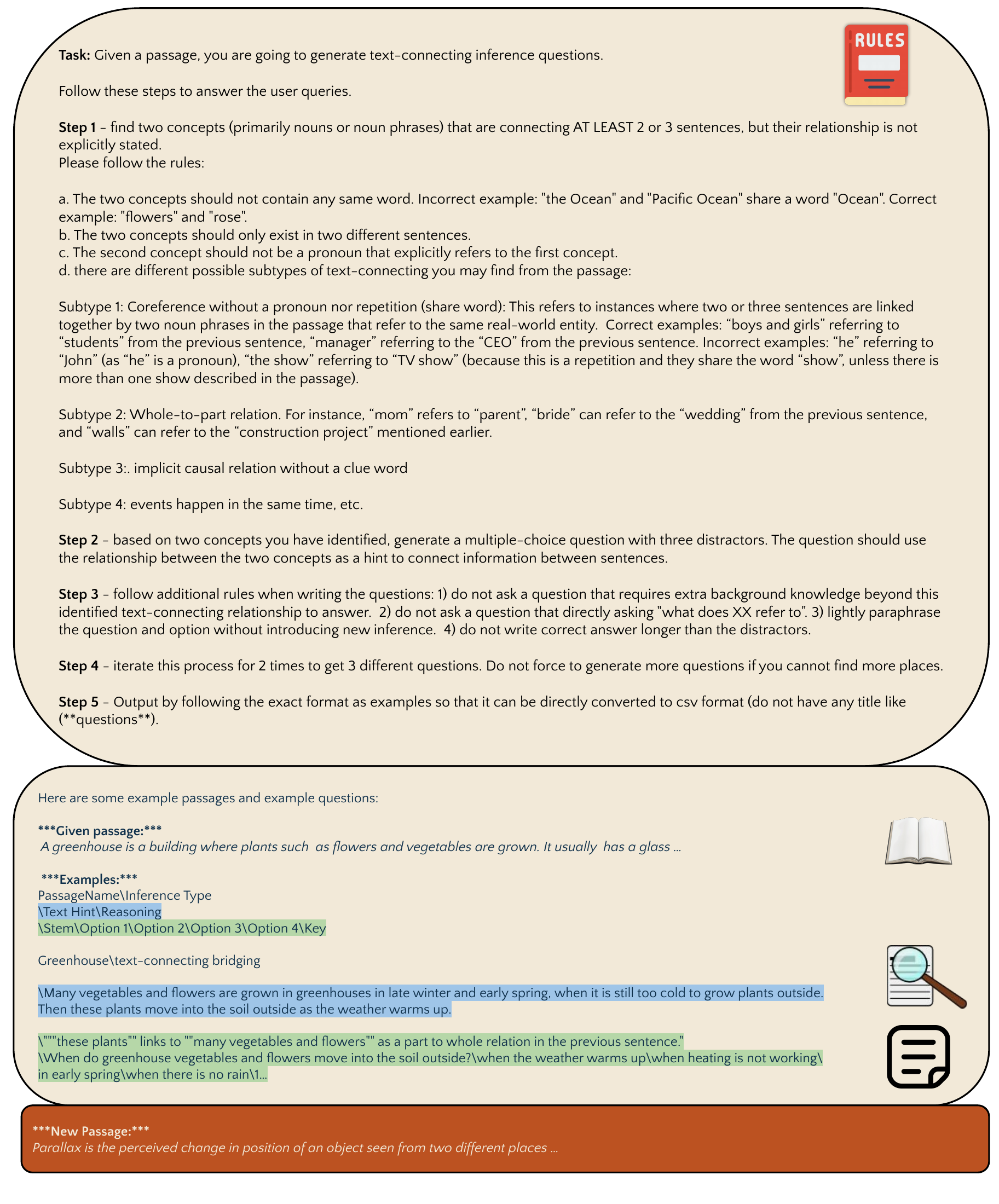}
  \caption{Few-shot prompting using Chain-of-Thought for generating text-connecting inference.}
  \label{fig:prompting_demo_text_connecting}
\end{figure*}

\begin{figure*}[h]
  \includegraphics[width=\linewidth]{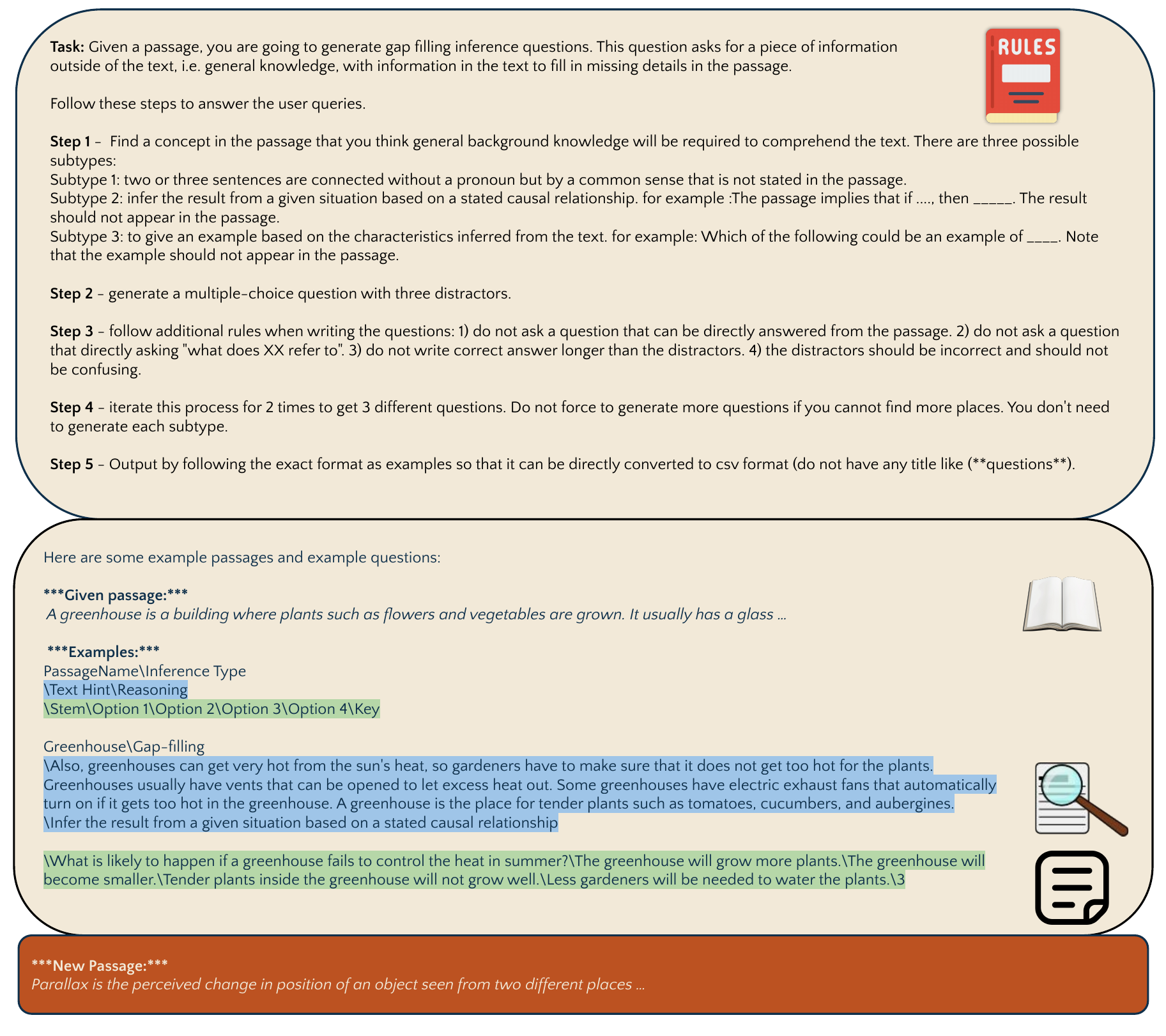}
  \caption{Few-shot prompting using Chain-of-Thought for generating gap-filling inference.}
  \label{fig:prompting_demo_gap_filling}
\end{figure*}

We present examples of our few-shot prompting design for pronominal bridging (Figure \ref{fig:prompting_demo}) text-connecting (Figure \ref{fig:prompting_demo_text_connecting}) and gap-filling (Figure \ref{fig:prompting_demo_gap_filling}) respectively. The rules are identical for both the standard and CoT prompts; the only difference is that CoT includes a text hint and reasoning in the training examples (see blue highlight in the figure). Accordingly, in the CoT condition, we expect the output to include a text hint and reasoning along with the generated questions. 

\end{document}